\documentclass{article}

\usepackage{iftex}
\ifPDFTeX
  \usepackage[utf8]{inputenc}
\fi
\usepackage[T1]{fontenc}

\usepackage{genai4health_2026}

\usepackage[hidelinks]{hyperref}
\usepackage{url}
\usepackage{booktabs}
\usepackage{amsfonts}
\usepackage{amsmath}
\usepackage{nicefrac}
\usepackage{microtype}
\usepackage{xcolor}
\usepackage{multirow}
\usepackage{graphicx}
\usepackage{float}

\usepackage{tikz}
\usetikzlibrary{shapes.geometric, arrows.meta, positioning, fit, backgrounds, calc}

\title{Bridging the EHR Divide: Asymmetric Contrastive Learning for Cross-National Medical Representation Transfer}

\author{%
  Qingyang Zhang \\
  Information Networking Institute \\
  Carnegie Mellon University \\
  Pittsburgh, United States \\
  \texttt{stevenz3@andrew.cmu.edu} \\
}
\begin{document}

\maketitle

\begin{abstract}

Cross-system transfer of longitudinal Electronic Health Record (EHR)
representations is challenging because clinical coding, patient populations,
and healthcare workflows differ substantially across institutions and
countries. We introduce \textbf{Asymmetric Supervised Contrastive Learning
(Asymmetric SupCon)}, a task-specific pre-training objective motivated by the
heterogeneity of negative clinical outcomes. The objective clusters patients
sharing a target positive outcome without explicitly attracting negative
trajectories toward one another.
We pre-train temporal Transformer encoders on longitudinal records from
3.98 million patients in the Taiwanese National Health Insurance Research
Database (NHIRD) and transfer them to two U.S. EHR datasets, MIMIC-IV and
EHRSHOT.
A hybrid semantic mapping pipeline combining direct mappings with
embedding-based retrieval enables transfer across heterogeneous clinical
vocabularies.
On MIMIC-IV, NHIRD pre-training consistently improves over random
initialization while substantially narrowing the performance gap to
task-specific in-domain pre-training. On EHRSHOT, the transferred
models show particularly strong few-shot performance for incident disease
prediction. A controlled objective ablation under a matched pre-training scale shows that
Asymmetric SupCon achieves higher mean AUPRC than direct supervised BCE
transfer on all four evaluated tasks and Standard SupCon on three of four,
with a 0.003 AUPRC deficit on readmission. These results support asymmetric contrastive
pre-training as an effective approach for task-specific cross-national EHR
representation transfer. Code is available at \url{https://github.com/qingYzhang/Asymmetric_SupCon}.

\end{abstract}

\section{Introduction}

Representation learning from longitudinal Electronic Health Records (EHRs)
has become an important approach to clinical prediction
\cite{rajkomar2018scalable,li2020behrt,rasmy2021medbert,steinberg2021clmbr}.
However, clinical forecasting presents an unusual representation-learning
problem: patients sharing a positive future outcome may exhibit common
prognostic factors, whereas the corresponding negative population can contain
highly heterogeneous clinical trajectories. Prior EHR phenotyping studies have
shown that constructing appropriate control populations can be sensitive to
heterogeneous healthcare utilization and incomplete case ascertainment
\cite{castro2014matched,zhang2020positiveunlabeled}.
Standard supervised contrastive learning \cite{khosla2020supervised} treats
both classes symmetrically and therefore explicitly attracts all examples
sharing the same binary label. This motivates our hypothesis that, for
heterogeneous clinical negatives, such attraction may unnecessarily compress
clinically meaningful variation.

To address this issue, we introduce a task-specific
\textbf{Asymmetric Supervised Contrastive (Asymmetric SupCon)} objective for
longitudinal clinical representation learning. The objective retains
positive-pair attraction while removing explicit attraction among negative
trajectories. We evaluate whether these outcome-specific risk representations,
learned from a large-scale Taiwanese claims database, can transfer to U.S.
clinical prediction tasks despite substantial geographic, institutional, and
coding-system shifts.

To evaluate this question, we pre-train temporal Transformer encoders using a
large Taiwanese NHIRD cohort and evaluate transfer on MIMIC-IV and EHRSHOT.
Because EHRSHOT contains clinical ontologies that do not directly match the
NHIRD vocabulary, we use a hybrid semantic alignment pipeline combining direct mappings with
embedding-based retrieval to enable cross-ontology evaluation.

Our contributions are:

\begin{itemize}
    \item \textbf{Asymmetric contrastive pre-training.}
    We introduce a task-specific contrastive objective that clusters trajectories
    sharing a positive future outcome without explicitly attracting the
    heterogeneous negative population.

    \item \textbf{Controlled objective evaluation.}
    Under a matched NHIRD pre-training scale, we compare Asymmetric SupCon with
    direct supervised BCE transfer, MLM, and Standard SupCon. Asymmetric SupCon achieves higher mean AUPRC than BCE transfer on all four
evaluated tasks and Standard SupCon on three of four.
    
    \item \textbf{Cross-national evaluation at scale.}
    We transfer representations pre-trained on 3.98 million Taiwanese NHIRD
    patients to MIMIC-IV and EHRSHOT, using semantic code alignment as an
    enabling component for cross-ontology evaluation.
\end{itemize}

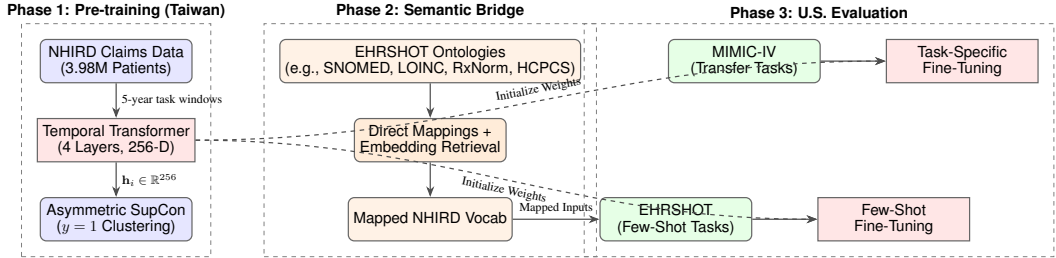
\begin{figure}[ht]
\centering
\resizebox{\textwidth}{!}{%
\begin{tikzpicture}[
    font=\sffamily,
    dataBox/.style={rectangle, rounded corners, draw=black!60, fill=blue!10, thick, minimum width=3.5cm, minimum height=1cm, align=center},
    modelBox/.style={rectangle, draw=black!60, fill=red!10, thick, minimum width=3.5cm, minimum height=1cm, align=center},
    mapBox/.style={rectangle, rounded corners, draw=black!60, fill=orange!15, thick, minimum width=3.5cm, minimum height=1cm, align=center},
    evalBox/.style={rectangle, draw=black!60, fill=green!10, thick, minimum width=3cm, minimum height=1cm, align=center},
    arrow/.style={-{Stealth[length=2.5mm]}, thick, draw=black!70},
    dashedArrow/.style={-{Stealth[length=2.5mm]}, dashed, thick, draw=black!70},
    groupNode/.style={rectangle, dashed, draw=black!50, inner sep=10pt}
]

\node[dataBox] (nhird)
{NHIRD Claims Data\\(3.98M Patients)};
\node[modelBox, below=0.8cm of nhird] (transformer) {Temporal Transformer\\(4 Layers, 256-D)};
\node[dataBox, below=0.8cm of transformer] (supcon) {Asymmetric SupCon\\($y=1$ Clustering)};

\draw[arrow] (nhird) -- node[right, font=\footnotesize] {5-year task windows} (transformer);
\draw[arrow] (transformer) -- node[right, font=\footnotesize] {$\mathbf{h}_i \in \mathbb{R}^{256}$} (supcon);

\node[groupNode, fit=(nhird)(transformer)(supcon), label=above:\textbf{Phase 1: Pre-training (Taiwan)}] (phase1) {};

\node[dataBox, right=2cm of nhird, fill=orange!10] (us_codes) {EHRSHOT Ontologies\\(e.g., SNOMED, LOINC, RxNorm, HCPCS)};
\node[mapBox, below=0.8cm of us_codes] (qwen)
{Direct Mappings +\\Embedding Retrieval};
\node[dataBox, below=0.8cm of qwen, fill=orange!10] (mapped)
{Mapped NHIRD Vocab};
\draw[arrow] (us_codes) -- (qwen);
\draw[arrow] (qwen) -- (mapped);

\node[groupNode, fit=(us_codes)(qwen)(mapped), label=above:\textbf{Phase 2: Semantic Bridge}] (phase2) {};

\node[dataBox, right=2cm of us_codes, fill=green!10] (mimic)
    {MIMIC-IV\\(Transfer Tasks)};
\node[dataBox, right=2cm of mapped, fill=green!10] (ehrshot)
    {EHRSHOT\\(Few-Shot Tasks)};

\node[modelBox, right=1.5cm of mimic] (mimicft)
    {Task-Specific\\Fine-Tuning};
\node[modelBox, right=1.5cm of ehrshot] (ehrft)
    {Few-Shot\\Fine-Tuning};

\draw[arrow] (mimic) -- (mimicft);
\draw[arrow] (ehrshot) -- (ehrft);

\draw[dashedArrow] (transformer.east) to[out=0, in=180]
    node[above, font=\footnotesize, sloped] {Initialize Weights}
    (mimicft.west);

\draw[dashedArrow] (transformer.east) to[out=0, in=180]
    node[below, font=\footnotesize, sloped] {Initialize Weights}
    (ehrft.west);

\draw[arrow] (mapped.east) --
    node[above, font=\footnotesize] {Mapped Inputs}
    (ehrshot.west);

\node[groupNode, fit=(mimic)(ehrshot)(mimicft)(ehrft),
      label=above:\textbf{Phase 3: U.S. Evaluation}] (phase3) {};

\end{tikzpicture}
}
\caption{
Overview of the cross-national transfer pipeline.
\textbf{Phase 1}: Task-specific temporal encoders are pre-trained on Taiwanese
NHIRD trajectories using Asymmetric SupCon.
\textbf{Phase 2}: Heterogeneous EHRSHOT clinical concepts are mapped to the NHIRD vocabulary
using direct mappings when available and embedding-based semantic retrieval
otherwise.
\textbf{Phase 3}: The transferred encoders are adapted to MIMIC-IV and EHRSHOT
prediction tasks, including few-shot EHRSHOT settings.
}
\label{fig:pipeline}
\end{figure}

\section{Related Work}

\paragraph{Longitudinal EHR Representation Learning.}
A growing body of work learns reusable patient representations from structured
longitudinal EHRs. BEHRT \cite{li2020behrt}, Med-BERT
\cite{rasmy2021medbert}, and CEHR-BERT \cite{pang2021cehr} adapt
Transformer-style pre-training to sequences of diagnoses, medications, and
clinical events. CLMBR \cite{steinberg2021clmbr} instead learns patient
representations through clinical language modeling and demonstrates improved
label efficiency for downstream prediction. More recently, MOTOR
\cite{steinberg2024motor} scales structured-record pre-training to large EHR
and claims corpora and evaluates transfer across multiple databases.
Complementary generative EHR models such as TransformEHR
\cite{yang2023transformehr} and Foresight \cite{kraljevic2024foresight}
directly model future clinical events or patient timelines.

Across these lines of work, pre-training is predominantly self-supervised or
generative, using objectives such as masked-token, next-event, future-event,
or time-to-event prediction. Our work instead studies task-specific supervised
representations designed explicitly around a target clinical outcome, with a
particular focus on whether such representations remain useful across national
healthcare systems and heterogeneous clinical vocabularies. Recent work has
also emphasized the importance of rigorous downstream evaluation when assessing
claims of general-purpose EHR foundation models
\cite{wornow2023shaky,wornow2023ehrshot}.

\paragraph{Contrastive Learning for Clinical Risk Prediction.}
Supervised contrastive learning \cite{khosla2020supervised} organizes
representations using class labels, and SCEHR \cite{zang2021scehr}
demonstrates its utility for longitudinal EHR risk prediction under class
imbalance. Our work focuses on a different property of binary clinical
forecasting: positive patients may share a specific future event while the
negative population can remain highly heterogeneous. We therefore remove
negative--negative attraction and use only eligible positive-class samples as
anchors. The term asymmetric supervised contrastive learning has also appeared
in continual learning, notably Co2L \cite{cha2021co2l}, where the asymmetry
distinguishes current-task anchors from replay-buffer examples. Our formulation
is distinct: the asymmetry is defined between positive and negative outcome
classes rather than between current and previously observed tasks.

\paragraph{Cross-Site Transfer and Clinical Vocabulary Alignment.}
Cross-site portability remains a central challenge for structured EHR models.
Recent studies have evaluated pretrained EHR representations across hospitals
and shown that external models can retain useful predictive and few-shot
performance after site transfer \cite{steinberg2024motor,guo2024multicenter}.
Common data models such as OMOP facilitate such interoperability by
standardizing schemas and clinical vocabularies \cite{voss2015omop}.
Complementary work on biomedical concept representations, including SapBERT
\cite{liu2021sapbert} and CODER \cite{yuan2022coder}, uses semantic embedding
spaces for medical entity alignment and normalization. Most recently, MedRep
\cite{kim2026medrep} explicitly targets EHR vocabulary portability using
standardized concept representations and evaluates external generalization on
MIMIC-IV and EHRSHOT.

Alternative approaches reduce dependence on code-specific schemas through
textual or standardized event representations. GenHPF
\cite{hur2024genhpf} converts heterogeneous EHR schemas and medical codes into
hierarchical textual representations for multi-source learning, while
MEDS-Torch \cite{oufattole2024medstorch} provides a standardized experimental
pipeline that compares multiple EHR tokenization strategies, including
text-based code representations, together with transfer-learning objectives.
Our setting differs in that we transfer outcome-specific representations
across national healthcare systems and map target concepts into a fixed
NHIRD-compatible vocabulary; semantic alignment serves as an enabling
component rather than the primary learning objective.

\section{Data and Methodology}

Figure~\ref{fig:pipeline} summarizes the framework. We first encode bounded
longitudinal patient histories with a temporal Transformer and pre-train a
task-specific encoder using Asymmetric SupCon. We then transfer the learned representations to U.S. datasets,
using semantic code alignment where the source and target vocabularies differ.

\subsection{Temporal Patient Representation and Bounding}

We model patient histories as discrete temporal sequences of medical tokens.
To focus the encoder on clinically relevant medium-to-long-term temporal
patterns while limiting distant historical information, we use at most five
years of longitudinal history ending at each task-specific prediction boundary.

Let a patient $i$ be represented by a longitudinal sequence of $T$ clinical
visits within this 5-year window,
$S_i=(V_{i,1},\dots,V_{i,T})$.

\paragraph{Clinical Tokenization.}
Each visit contains a variable-length set of diagnosis, procedure, and
medication codes. Codes are mapped to a unified NHIRD vocabulary and serialized
chronologically across visits before sequence truncation and padding. Patient
trajectories are truncated to a maximum context window of 2,048 tokens.

We employ a Transformer-based encoder architecture configured with 4 hidden layers, 8 attention heads, and a hidden dimension of 256. The input integrates learned token embeddings with sequential positional embeddings. After processing the temporal sequence through the Transformer, we apply mean pooling across all valid (non-padded) tokens and $\ell_2$-normalize the output to generate the final continuous patient representation $\mathbf{h}_i \in \mathbb{R}^{256}$.

\subsection{Asymmetric Supervised Contrastive Pre-training}
\label{sec:asym_supcon}

The primary methodological contribution of this work is an asymmetric
supervised contrastive objective designed for task-specific longitudinal
clinical forecasting. For each target clinical outcome, we pre-train a
separate temporal encoder. Let $y_i\in\{0,1\}$ indicate whether patient $i$
experiences the target outcome under its task-specific outcome definition.
The resulting representation is therefore task specific rather than a single
universal representation shared across arbitrary downstream tasks.

\paragraph{Task-specific outcome labels.}
For each target outcome, we construct a corresponding NHIRD pre-training label
using a claims-based operationalization of the clinical endpoint and, when
applicable, its prediction horizon. For Acute MI, lupus, hyperlipidemia, and
hypertension, incident disease labels are defined using future first-occurrence
diagnoses while excluding prior evidence of the target condition. Pancreatic
cancer uses a dedicated matched cancer-screening construction described in
Appendix~\ref{appendix:task_definitions}. Readmission uses a 30-day interval to
the next inpatient admission, whereas mortality is defined using the subsequent
eligible hospitalization within a 90-day window. Long LOS and ICU-related
outcomes follow their task-specific hospitalization definitions.

Standard supervised contrastive learning encourages examples sharing a class
label to occupy nearby regions of the representation space. In clinical
forecasting, however, the semantic structure of the two classes can be highly
asymmetric. Positive patients share a specific future outcome, whereas negative
patients may differ substantially in diagnoses, treatments, disease severity,
and utilization history.

We therefore hypothesize that explicitly attracting all negative examples
toward one another may over-compress meaningful variation within the baseline
population. To avoid this constraint, Asymmetric SupCon applies positive-pair
attraction only to patients sharing the target positive outcome.

In the Standard SupCon baseline, every sample acts as an anchor and its
positive set consists of all other samples in the batch sharing the same class
label. Consequently, negative-class samples are explicitly attracted to one
another when used as same-label pairs.

Let $\mathcal{I}$ denote a training batch and define

\[
\mathcal{P}(i)
=
\{p\in\mathcal{I}\setminus\{i\}:y_i=y_p=1\},
\qquad
\mathcal{I}_{+}
=
\{i\in\mathcal{I}:y_i=1,\ |\mathcal{P}(i)|>0\}.
\]

For $\ell_2$-normalized patient representations $\mathbf{h}_i$, we optimize

\[
\mathcal{L}_{\mathrm{Asym}}
=
-\frac{1}{|\mathcal{I}_{+}|}
\sum_{i\in\mathcal{I}_{+}}
\frac{1}{|\mathcal{P}(i)|}
\sum_{p\in\mathcal{P}(i)}
\log
\frac{
\exp(\mathbf{h}_i^\top\mathbf{h}_p/\tau)
}{
\sum_{a\in\mathcal{I}\setminus\{i\}}
\exp(\mathbf{h}_i^\top\mathbf{h}_a/\tau)
},
\]

where $\tau=0.07$ is the temperature parameter.
Positive anchors without another positive example in the batch are omitted
from the contrastive loss.

Under this objective, positive trajectories sharing the future outcome are
explicitly encouraged to become similar. Negative examples remain in the contrastive denominator and therefore
influence the relative separation of positive representations, but no
negative--negative attractive term is introduced. We do not assume that
Standard SupCon necessarily collapses the negative class; rather, we
hypothesize that removing this attraction is beneficial when negative clinical
trajectories are heterogeneous. This hypothesis is evaluated through the
controlled objective ablation in Section~\ref{sec:objective_ablation}.

\subsection{Massive-Scale Pre-training via NHIRD}
We pre-train on a large longitudinal subset of the Taiwanese National Health
Insurance Research Database (NHIRD) \cite{lin2018nhird}, containing
3,989,369 patients and records drawn from 1996--2013. The source corpus contains up to 15 years of longitudinal
records per patient, while each task-specific training example uses at most the
five years ending at its prediction boundary. The corpus includes approximately
1.44 billion inpatient, outpatient, and pharmacy visits and 7.92 billion
diagnosis, procedure, and medication tokens represented in the NHIRD vocabulary.

Pre-training uses a strict patient-level split. We assign 80\% of the full
cohort (3,191,495 patients) to the training partition and the remaining 20\%
to the validation partition. Task-specific eligibility criteria are applied
within each partition, and validation patients are never used for gradient
updates.

\subsection{Semantic Alignment for Cross-Domain Evaluation}
\label{sec:semantic_mapping}

Cross-national transfer introduces mismatches between the clinical vocabularies
used by NHIRD and the U.S. target datasets. MIMIC-IV is harmonized to the NHIRD-compatible vocabulary
using direct code mappings available in our preprocessing resources where
needed, whereas EHRSHOT contains a broader mixture of heterogeneous
ontologies, including SNOMED, LOINC, RxNorm, HCPCS, and related coding systems.

For EHRSHOT, we therefore use a hybrid code-alignment pipeline. When a direct
mapping to the NHIRD-compatible vocabulary is available in our preprocessing
resources, we retain it. Otherwise, we formulate code alignment as a dense
semantic retrieval problem: source concept descriptions are embedded using
Qwen3-Embedding-8B \cite{zhang2025qwen3embedding}, and nearest-neighbor
retrieval is performed against the corresponding NHIRD vocabulary using
FAISS \cite{johnson2019billion}.

For embedding-derived mappings, we retain candidates whose cosine similarity
exceeds $\tau_{\mathrm{sim}}=0.8$. This operating point was selected
empirically to balance the breadth of the mapped NHIRD vocabulary and
target-side mapping collisions. Additional comparisons across embedding models
and similarity thresholds are reported in
Appendix~\ref{appendix:mapping}.

We emphasize that semantic alignment is an enabling component rather than the
primary methodological contribution of this work. Its role is to make
cross-ontology evaluation possible and thereby test whether task-specific
representations learned with Asymmetric SupCon remain useful under substantial
cross-system and coding-system shifts.

\subsection{Downstream Formulation and Adaptation}
\label{sec:downstream}

For each task, the model observes at most five years of longitudinal history
ending at the corresponding task-specific prediction index. Information after
the prediction boundary is excluded from the model input. Readmission uses a 30-day interval to the next inpatient admission, mortality
uses a 90-day outcome window, and incident disease tasks use one-year
prediction horizons. Long LOS and ICU-related outcomes follow their
task-specific hospitalization definitions. NHIRD pre-training cohort and label construction procedures are provided in
Appendix~\ref{appendix:task_definitions}.

For MIMIC-IV, patients are partitioned at the patient level into mutually
exclusive 80/10/10 training, validation, and test sets. We compare
(1) random initialization, (2) task-specific in-domain MIMIC pre-training, and
(3) task-specific cross-domain NHIRD pre-training. The in-domain baseline uses the same temporal encoder architecture and
Asymmetric SupCon objective as the cross-domain model, but is pre-trained
exclusively on patients from the MIMIC-IV training cohort. This 80\% training
cohort is further partitioned internally for pre-training optimization and
checkpoint selection; the outer validation and test cohorts remain untouched. The same downstream architecture and model-selection procedure
are used across all three settings. Hyperparameters are selected exclusively
on the validation split, and final performance is reported on the held-out
test split.

For EHRSHOT, we follow the benchmark's few-shot evaluation setting and adapt
the corresponding NHIRD-pretrained encoder using the available labeled examples
for each target task. We evaluate multiple $k$-shot regimes together with the
full-data setting, select downstream hyperparameters using the predefined
validation data, and report AUPRC on the held-out test set.

\section{Downstream Tasks and Evaluation Setup}

To rigorously evaluate the transferability of our NHIRD-pretrained encoders,
we benchmark performance on two U.S.-based evaluation resources:
MIMIC-IV \cite{johnson2023mimic} and the EHRSHOT benchmark
\cite{wornow2023ehrshot}.

\subsection{Task Definitions}
\begin{itemize}
    \item \textbf{MIMIC-IV Tasks:}
    We evaluate three binary clinical prediction tasks:
    \textbf{30-Day Readmission}, \textbf{90-Day Mortality}, and
    \textbf{Long Length-of-Stay (Long LOS)} prediction.
    The same outcome definitions, prediction horizons, and exclusion criteria
    used to construct the corresponding NHIRD pre-training tasks are applied to
    MIMIC-IV; these definitions are detailed in
    Appendix~\ref{appendix:task_definitions}. The MIMIC-IV cohort includes all eligible inpatient hospitalizations,
regardless of ICU involvement, and is not restricted to ICU stays.
    \item \textbf{EHRSHOT Tasks:}
For EHRSHOT, heterogeneous input concepts are aligned to the NHIRD vocabulary
through our semantic mapping pipeline. We evaluate three operational outcomes:
\textbf{Long LOS} ($\geq 7$ days), \textbf{ICU Admission}, and
\textbf{30-Day Readmission}; and five one-year incident diagnosis tasks:
\textbf{Acute MI}, \textbf{Pancreatic Cancer}, \textbf{Lupus},
\textbf{Hyperlipidemia}, and \textbf{Hypertension}.
\end{itemize}

\subsection{Evaluation Metrics}

Because several clinical outcomes are highly imbalanced, AUPRC is used as the
primary evaluation metric. For MIMIC-IV, we additionally report AUROC and
F1-score. EHRSHOT results are reported using AUPRC to follow the benchmark's
few-shot evaluation protocol.
\section{Results}

\subsection{Cross-National Transfer on MIMIC-IV}

\begin{table}[H]
\centering
\caption{
Downstream performance on MIMIC-IV. Results are reported as mean $\pm$ standard deviation over five downstream
random seeds.
All models use the same downstream architecture, data split, and
hyperparameter-selection protocol.
The in-domain and cross-domain models are initialized from task-specific
pre-trained encoders, while the train-from-scratch model is randomly initialized.
Best mean results are bolded and second-best results are underlined.
}
\label{tab:mimic_results}
\resizebox{\textwidth}{!}{
\begin{tabular}{llccc}
\toprule
\textbf{Task}
& \textbf{Metric}
& \textbf{Train-from-Scratch}
& \textbf{In-Domain (MIMIC $\rightarrow$ MIMIC)}
& \textbf{Cross-Domain (NHIRD $\rightarrow$ MIMIC)} \\
\midrule

\multirow{3}{*}{30-Day Readmission}
& AUROC
& $0.6252 \pm 0.0031$
& $\mathbf{0.6372 \pm 0.0024}$
& $\underline{0.6331 \pm 0.0027}$ \\

& AUPRC
& $0.4127 \pm 0.0038$
& $\mathbf{0.4205 \pm 0.0030}$
& $\underline{0.4195 \pm 0.0032}$ \\

& F1
& $0.4665 \pm 0.0042$
& $\mathbf{0.4734 \pm 0.0036}$
& $\underline{0.4723 \pm 0.0039}$ \\

\midrule

\multirow{3}{*}{90-Day Mortality}
& AUROC
& $0.8883 \pm 0.0028$
& $\mathbf{0.9237 \pm 0.0021}$
& $\underline{0.9142 \pm 0.0025}$ \\

& AUPRC
& $0.5029 \pm 0.0052$
& $\mathbf{0.5933 \pm 0.0041}$
& $\underline{0.5693 \pm 0.0048}$ \\

& F1
& $0.4799 \pm 0.0049$
& $\mathbf{0.5201 \pm 0.0040}$
& $\underline{0.5094 \pm 0.0045}$ \\

\midrule

\multirow{3}{*}{Long LOS ($\geq 7$ Days)}
& AUROC
& $0.7848 \pm 0.0035$
& $\mathbf{0.8554 \pm 0.0028}$
& $\underline{0.8330 \pm 0.0031}$ \\

& AUPRC
& $0.5536 \pm 0.0047$
& $\mathbf{0.6825 \pm 0.0039}$
& $\underline{0.6378 \pm 0.0044}$ \\

& F1
& $0.5057 \pm 0.0046$
& $\mathbf{0.5867 \pm 0.0038}$
& $\underline{0.5592 \pm 0.0042}$ \\

\bottomrule
\end{tabular}
}
\end{table}

Table~\ref{tab:mimic_results} shows that NHIRD pre-training consistently
improves over random initialization while remaining below task-specific
in-domain pre-training. For 30-day readmission, cross-domain AUPRC reaches 0.4195 versus
0.4205 for in-domain pre-training and 0.4127 from scratch. For mortality,
cross-domain pre-training improves AUPRC from 0.5029 to 0.5693, compared with
0.5933 in-domain; for Long LOS, it improves AUPRC from 0.5536 to 0.6378, compared
with 0.6825 in-domain.

These results indicate that outcome-specific clinical risk representations can
retain substantial predictive value after transfer across healthcare systems,
although the remaining gap for mortality and Long LOS shows that transferability is
task dependent.

\subsection{Few-Shot Generalization on EHRSHOT}

\begin{table*}[t]
\centering
\caption{
EHRSHOT few-shot performance (AUPRC).
Our results are averaged over three downstream random seeds for each setting.
``Best Baseline'' is the best published result among CLMBR, GBM,
Logistic Regression, and Random Forest for each setting
\cite{wornow2023ehrshot}. Full baseline results across all $k$ are provided
in Appendix Table~\ref{tab:ehrshot_full}. Bold indicates the better result
between the best baseline and our task-specific NHIRD-pretrained model.
}
\label{tab:ehrshot_results}
\small
\setlength{\tabcolsep}{3.2pt}
\resizebox{\textwidth}{!}{
\begin{tabular}{llcccccc}
\toprule
\textbf{Task} & \textbf{Model}
& \textbf{All}
& $\mathbf{k=1}$
& $\mathbf{k=8}$
& $\mathbf{k=16}$
& $\mathbf{k=64}$
& $\mathbf{k=128}$ \\
\midrule

\multirow{2}{*}{Long LOS}
& Best Baseline & \textbf{0.576} & 0.282 & 0.315 & 0.337 & \textbf{0.439} & \textbf{0.473} \\
& AsymSupCon & 0.517 & \textbf{0.298} & \textbf{0.333} & \textbf{0.355} & 0.431 & 0.450 \\
\midrule

\multirow{2}{*}{ICU Admission}
& Best Baseline & \textbf{0.324} & \textbf{0.093} & \textbf{0.142} & \textbf{0.181} & \textbf{0.209} & \textbf{0.256} \\
& AsymSupCon & 0.138 & 0.055 & 0.080 & 0.090 & 0.108 & 0.129 \\
\midrule

\multirow{2}{*}{30-Day Readmission}
& Best Baseline & \textbf{0.412} & 0.150 & 0.237 & \textbf{0.297} & \textbf{0.368} & \textbf{0.350} \\
& AsymSupCon & 0.343 & \textbf{0.172} & \textbf{0.238} & 0.243 & 0.278 & 0.289 \\
\midrule

\multirow{2}{*}{Acute MI}
& Best Baseline & 0.184 & 0.086 & 0.096 & 0.118 & 0.154 & 0.158 \\
& AsymSupCon & \textbf{0.191} & \textbf{0.106} & \textbf{0.135} & \textbf{0.153} & \textbf{0.178} & \textbf{0.178} \\
\midrule

\multirow{2}{*}{Pancreatic Cancer}
& Best Baseline & \textbf{0.472} & 0.058 & 0.080 & 0.123 & 0.181 & \textbf{0.243} \\
& AsymSupCon & 0.305 & \textbf{0.074} & \textbf{0.095} & \textbf{0.151} & \textbf{0.187} & 0.229 \\
\midrule

\multirow{2}{*}{Lupus}
& Best Baseline & \textbf{0.179} & 0.012 & 0.015 & 0.017 & 0.061 & 0.079 \\
& AsymSupCon & 0.093 & \textbf{0.022} & \textbf{0.024} & \textbf{0.036} & \textbf{0.085} & \textbf{0.084} \\
\midrule

\multirow{2}{*}{Hyperlipidemia}
& Best Baseline & 0.258 & 0.144 & 0.180 & 0.183 & 0.210 & 0.244 \\
& AsymSupCon & \textbf{0.297} & \textbf{0.146} & \textbf{0.207} & \textbf{0.234} & \textbf{0.248} & \textbf{0.265} \\
\midrule

\multirow{2}{*}{Hypertension}
& Best Baseline & 0.258 & \textbf{0.151} & 0.205 & 0.224 & 0.233 & 0.243 \\
& AsymSupCon & \textbf{0.295} & 0.139 & \textbf{0.254} & \textbf{0.266} & \textbf{0.279} & \textbf{0.280} \\

\bottomrule
\end{tabular}
}
\end{table*}

Our model additionally leverages task-specific source-domain supervision from
NHIRD before EHRSHOT adaptation. The comparison therefore evaluates
target-domain label efficiency under cross-national transfer rather than
equalizing source-domain supervision across methods.

Table~\ref{tab:ehrshot_results} shows a clear task-dependent transfer pattern.
At $k=16$, AsymSupCon exceeds the strongest published baseline on all five
incident disease tasks: Acute MI, Pancreatic Cancer, Lupus, Hyperlipidemia,
and Hypertension. Similar gains appear across several lower-shot regimes,
indicating strong label efficiency for disease-risk transfer. The full $k$-shot sweep in Appendix Table~\ref{tab:ehrshot_full} further
shows that the value of cross-national pre-training is task- and
label-regime-dependent. Several incident-disease tasks exhibit their clearest
gains in low-to-moderate shot settings, while the advantage narrows or
reverses for some tasks as substantially more target-domain labels become
available. This pattern is consistent with cross-national pre-training being
most useful in label-limited target settings, although the comparison does not
equalize source-domain supervision across methods.

Transfer is weaker for operational outcomes. Our model is competitive with
published baselines on Long LOS and selected low-shot readmission settings, but
substantially underperforms CLMBR on ICU Admission. This suggests that
disease-related clinical risk signals may transfer more reliably across healthcare
systems than outcomes strongly influenced by local admission, discharge, and
resource-allocation practices.

\subsection{Ablation Study: Effect of the Pre-training Objective}
\label{sec:objective_ablation}

To separate the effect of task-specific source-domain supervision from the
effect of the pre-training objective, we conduct a controlled ablation using
approximately 400K NHIRD patients, corresponding to roughly 10\% of the full
3.98M-patient cohort, chosen for computational tractability.

All pre-trained variants use the same patient cohort, temporal encoder,
number of pre-training epochs, batch size, optimizer settings, and downstream
adaptation procedure. Objective-specific output heads differ as required by
the corresponding pre-training loss. We compare direct supervised transfer
using binary cross-entropy (BCE), Masked Language Modeling (MLM), Standard
Supervised Contrastive Learning, and the proposed Asymmetric SupCon objective.
A randomly initialized model without NHIRD pre-training is included as an
additional reference.

As shown in Table~\ref{tab:objective_ablation}, the supervised BCE
source-transfer baseline achieves AUPRCs of 0.414, 0.510, 0.116, and 0.185
for 30-day readmission, 90-day mortality, Acute MI, and Hyperlipidemia,
respectively. These results represent modest mean improvements over random
initialization across all four tasks, suggesting that task-specific
source-domain supervision itself can contribute to transfer.

Asymmetric SupCon nevertheless achieves higher mean AUPRC than direct
supervised BCE transfer on all four tasks, improving AUPRC by 0.003, 0.028, 0.028, and 0.043,
respectively. Under this matched 400K-patient pre-training setup, this suggests
that exposure to source-domain labels alone does not fully account for the
transfer gains obtained with the asymmetric contrastive objective.

Relative to Standard SupCon, Asymmetric SupCon achieves higher mean AUPRC on
90-day mortality ($0.517 \rightarrow 0.538$), Acute MI
($0.132 \rightarrow 0.144$), and Hyperlipidemia
($0.205 \rightarrow 0.228$). For 30-day readmission, Standard SupCon remains
slightly higher (0.420 versus 0.417), with a mean difference of 0.003 relative
to across-seed standard deviations of 0.004 for both models.

Asymmetric SupCon also achieves higher mean AUPRC than MLM on all four
evaluated tasks, whereas Standard SupCon does not do so uniformly; for
90-day mortality, MLM achieves 0.519 AUPRC compared with 0.517 for
Standard SupCon.

Together, these results suggest that both task-specific source supervision and
the choice of representation-learning objective influence cross-domain
transfer, with Asymmetric SupCon providing the strongest overall performance
on three of the four evaluated tasks.

\begin{table*}[t]
\centering
\caption{
Ablation of pre-training objectives using approximately 400K NHIRD patients
($\sim$10\% of the full cohort).
The four pre-trained variants use the same temporal encoder, NHIRD
pre-training cohort, number of training epochs, batch size, optimizer settings,
and downstream adaptation protocol. Objective-specific output heads differ as
required by the corresponding pre-training loss.
For task-specific supervised objectives (BCE, Standard SupCon, and Asymmetric
SupCon), a separate encoder is pre-trained for each target outcome.
The random-initialization baseline skips NHIRD pre-training while retaining
the same downstream architecture and adaptation protocol.
Results are reported as mean $\pm$ standard deviation over five downstream
random seeds.
MIMIC-IV results use the full downstream training set, while EHRSHOT results
use the $k=16$ few-shot setting.
Performance is measured by AUPRC.
}
\label{tab:objective_ablation}
\resizebox{\linewidth}{!}{
\begin{tabular}{lcccc}
\toprule
&
\multicolumn{2}{c}{\textbf{MIMIC-IV}}
&
\multicolumn{2}{c}{\textbf{EHRSHOT ($k=16$)}} \\
\cmidrule(lr){2-3}
\cmidrule(lr){4-5}

\textbf{Pre-training Objective}
& \textbf{30-Day Readmission}
& \textbf{90-Day Mortality}
& \textbf{Acute MI}
& \textbf{Hyperlipidemia} \\
\midrule

Random Initialization
& $0.413 \pm 0.004$
& $0.503 \pm 0.005$
& $0.112 \pm 0.008$
& $0.172 \pm 0.006$ \\

Supervised BCE Transfer
& $0.414 \pm 0.004$
& $0.510 \pm 0.005$
& $0.116 \pm 0.007$
& $0.185 \pm 0.006$ \\

Masked Language Modeling (MLM)
& $0.414 \pm 0.002$
& $\underline{0.519 \pm 0.004}$
& $0.121 \pm 0.007$
& $0.194 \pm 0.006$ \\

Standard SupCon
& $\mathbf{0.420 \pm 0.004}$
& $0.517 \pm 0.006$
& $\underline{0.132 \pm 0.006}$
& $\underline{0.205 \pm 0.004}$ \\

\textbf{Asymmetric SupCon (Ours)}
& $\underline{0.417 \pm 0.004}$
& $\mathbf{0.538 \pm 0.005}$
& $\mathbf{0.144 \pm 0.006}$
& $\mathbf{0.228 \pm 0.005}$ \\

\bottomrule
\end{tabular}
}
\end{table*}

\section{Discussion}

\subsection{Asymmetric Contrastive Learning}

The controlled ablation helps distinguish the effect of task-specific
source-domain supervision from the effect of the representation-learning
objective. Direct supervised BCE transfer provides modest mean
improvements over random initialization, suggesting that access to NHIRD
outcome labels is itself beneficial. However, Asymmetric SupCon achieves higher mean AUPRC than the BCE transfer
baseline on all four evaluated tasks under the same 400K-patient pre-training
scale.

Relative to Standard SupCon, Asymmetric SupCon achieves the strongest
performance on three of four tasks and is slightly lower on readmission.
These results do not establish that symmetric SupCon universally causes
representation collapse; rather, they suggest that both source supervision
and the structure of the pre-training objective matter, and that removing
explicit negative--negative attraction can improve task-specific clinical
transfer when the negative population is heterogeneous.

\subsection{Cross-National Transfer}

NHIRD-pretrained representations retain substantial predictive value on U.S.
datasets despite differences in coding systems, populations, and healthcare
practice. The EHRSHOT results further suggest that transferability depends on
the nature of the target outcome. Acute MI, Pancreatic Cancer, Lupus,
Hyperlipidemia, and Hypertension show strong few-shot transfer, whereas ICU
admission and readmission appear more sensitive to institution-specific care
patterns.

One plausible explanation is that incident disease prediction depends more
strongly on pathophysiologic risk signals that may be shared across
populations, whereas operational outcomes are more directly shaped by local
care pathways, admission and discharge practices, and resource availability.
This interpretation is consistent with the observed transfer pattern, although
our experiments do not directly isolate the mechanisms responsible for these
differences.

Importantly, our pre-training is task specific: a separate encoder is learned
for each target outcome. The results therefore demonstrate transfer of
outcome-specific risk representations rather than a single universal clinical
foundation representation.

\subsection{Limitations}

Semantic code mapping enables EHRSHOT transfer but introduces information loss
and possible alignment ambiguity. In particular, multiple external concepts
may converge on the same NHIRD target concept, while some source concepts may
produce mappings across multiple target code types. The collision analysis
characterizes this structural ambiguity but does not establish semantic mapping
correctness.

Comparisons with published EHRSHOT baselines do not equalize source-domain
supervision: our model additionally uses task-specific NHIRD labels before
target-domain adaptation. These experiments should therefore be interpreted as
evaluations of target-domain label efficiency under cross-national transfer
rather than source-supervision-matched comparisons. The supervised BCE
source-transfer baseline in Section~\ref{sec:objective_ablation} provides a
controlled comparison for isolating the effect of the pre-training objective,
but this analysis is limited to a 400K-patient NHIRD subset and four
representative tasks. Whether the relative ranking of objectives persists
across the full task set and full-scale NHIRD pre-training remains to be
established.

Our method is motivated by the hypothesis that negative clinical populations
are more heterogeneous than patients sharing a specific positive outcome.
Although the controlled objective ablation evaluates the downstream
consequences of removing negative--negative attraction, we do not directly
quantify within-class clinical or representation heterogeneity. Further work
should test this mechanism explicitly.

The reported standard deviations reflect downstream random variation for fixed
pre-trained checkpoints rather than uncertainty across independent large-scale
pre-training runs. In addition, our cross-national evaluation is
unidirectional, using Taiwan as the source domain and U.S. datasets as target
domains. Evaluation across additional national healthcare systems and reverse
transfer directions is needed to establish broader geographic generalizability.

Finally, we do not report subgroup-specific discrimination or calibration
across demographic and clinical populations. Such analyses are particularly
important for clinical prediction models and are emphasized by TRIPOD+AI
reporting guidance \cite{collins2024tripodai}. Future work should evaluate
subgroup performance, calibration, and target-data scaling under
source-supervision-matched settings.

\section{Conclusion}
We introduce Asymmetric SupCon, a task-specific contrastive pre-training
objective that clusters trajectories sharing a positive clinical outcome
without explicitly attracting heterogeneous negative examples.
Representations pre-trained on the Taiwanese NHIRD cohort consistently improve
over random initialization on MIMIC-IV and show strong few-shot
disease-prediction performance on EHRSHOT. Controlled comparisons with direct
supervised BCE transfer, MLM, and Standard SupCon further suggest that the
asymmetric objective contributes beyond task-specific source-label exposure
under a matched pre-training scale. Overall, the results demonstrate the
potential of outcome-specific contrastive pre-training for cross-national
longitudinal EHR transfer while highlighting important limitations for
healthcare-system-dependent outcomes.

\begin{ack}
We express our sincere gratitude to Prof. Chenyan Xiong for his invaluable guidance and support throughout this project. We also thank Liwen Sun for helpful discussions and assistance. Additionally, computing resources were generously provided by Carnegie Mellon University.
\end{ack}

\newpage
\bibliographystyle{plain}
\bibliography{references}

\newpage
\appendix

\section{Data Preprocessing and Task Formulation}
\label{appendix:preprocessing}

To ensure reproducibility and prevent data leakage, we apply task-specific
data cleaning and temporal boundary enforcement during NHIRD pre-training and
downstream evaluation. Semantic code alignment is additionally applied when
U.S. source vocabularies do not directly match the NHIRD vocabulary.

\subsection{NHIRD Pre-training Cohort and Label Definitions}
\label{appendix:task_definitions}
For each clinical outcome, we construct a separate task-specific dataset from
the longitudinal NHIRD records. Unless otherwise specified, the model input contains at most five years of
clinical history ending at the corresponding task-specific prediction
boundary. Information occurring after that boundary is excluded from the input.

\paragraph{30-Day Readmission.}
Each eligible inpatient hospitalization is paired with the patient's next
observed inpatient hospitalization. Let $t_i$ denote the admission date of the
index hospitalization and $t_{i+1}$ the admission date of the subsequent
hospitalization. The binary label is

\[
y_i =
\begin{cases}
1, & 1 \leq t_{i+1}-t_i \leq 30 \text{ days},\\
0, & t_{i+1}-t_i > 30 \text{ days}.
\end{cases}
\]

Same-day consecutive admissions ($t_{i+1}-t_i=0$) are excluded. Patients with
fewer than two eligible inpatient admissions do not contribute a readmission
example. Because a subsequent hospitalization is required to construct the
label, the final recorded inpatient hospitalization for each patient is not
used as an index encounter.

\paragraph{90-Day Mortality.}
For each eligible index hospitalization, we evaluate whether the immediately
subsequent eligible inpatient hospitalization carries a mortality indicator
within 90 days of the index admission. The label is

\[
y_i =
\mathbf{1}
\left[
m_{i+1}=1
\land
0 \le t_{i+1}-t_i \le 90 \text{ days}
\right],
\]

where $m_{i+1}$ denotes the mortality indicator associated with the subsequent
hospitalization. Otherwise, the index hospitalization is assigned a negative
label. Mortality indicators are derived from the available discharge-transfer
and mortality information during preprocessing.

\paragraph{Long Length of Stay (Long LOS).}
Length of stay is calculated from the admission and discharge dates of the
index hospitalization. When these timestamps cannot be used directly, the
available bed-day fields are used as a fallback. Hospitalizations with
$\mathrm{LOS}<1$ day are excluded. For all remaining encounters,

\[
y_i =
\begin{cases}
1, & \mathrm{LOS}_i \geq 7 \text{ days},\\
0, & 1 \leq \mathrm{LOS}_i < 7 \text{ days}.
\end{cases}
\]

\paragraph{ICU Admission.}
ICU involvement is identified from predefined ICU-stay order codes associated
with the index hospitalization. The binary label is

\[
y_i =
\mathbf{1}
\left[
\text{the index hospitalization contains an ICU-stay indicator}
\right].
\]

ICU-positive hospitalizations with $\mathrm{LOS}<1$ day are excluded.
All remaining eligible hospitalizations without an ICU-stay indicator are
treated as negative examples.

\paragraph{One-Year Incident Disease Prediction.}
For Acute MI, lupus, hyperlipidemia, and hypertension, we construct
post-index disease-prediction tasks from eligible inpatient hospitalizations.
An index hospitalization is eligible only when the target disease has not been
documented in the patient's history up to and including the index encounter.
A positive label is assigned when the target disease appears in a subsequent
visit within the following one-year prediction window. Patients with no
documented occurrence of the target disease provide negative examples when no
target diagnosis appears during the corresponding one-year future window.

The target diagnoses are identified using the following ICD-9 patterns:
Acute MI, \texttt{410*}; hypertension, \texttt{401*}; hyperlipidemia,
\texttt{272*}; and lupus, \texttt{7100}.

\paragraph{Pancreatic Cancer Prediction.}
Pancreatic cancer is constructed using a dedicated cancer-screening pipeline
and is identified by ICD-9 codes matching \texttt{157*}. For positive cases,
the first documented pancreatic-cancer diagnosis defines the target event.
Model input is constructed exclusively from clinical visits preceding this
diagnosis; the target diagnosis encounter itself is excluded. We use the
configured one-year pre-diagnosis window to construct the prediction history.

Negative examples are drawn from patients with no documented cancer diagnosis
in their longitudinal record. Control patients are matched to cancer cases
using demographic and longitudinal-history characteristics, including sex,
age, accumulated history at the target time, and total available history
length.

With the risk filter disabled in the main experiments, pancreatic-cancer cases
may include patients with prior non-pancreatic malignancies, whereas matched
controls have no documented cancer diagnosis.

\section{Implementation Details}
\label{appendix:implementation}

\paragraph{Encoder.}
All models trained in our pipeline use a four-layer temporal Transformer with eight attention
heads, hidden dimension 256, and feed-forward dimension 1,024. Clinical token
embeddings are combined with learned positional embeddings. Patient
trajectories are truncated to at most 2,048 tokens and dynamically padded
within each batch. Padding tokens are masked during self-attention and excluded
from representation pooling. The final patient representation is obtained by
mean-pooling valid token representations followed by $\ell_2$ normalization.

\paragraph{Asymmetric SupCon Pre-training.}
Unless otherwise specified, task-specific encoders are pre-trained for
10 epochs with batch size 128 using AdamW, a learning rate of
$1\times10^{-4}$, weight decay $1\times10^{-4}$, and contrastive
temperature $\tau=0.07$. Training batches are randomly shuffled and incomplete
final batches are discarded. Validation loss is evaluated after every epoch,
and the checkpoint with the lowest validation contrastive loss is retained.
The contrastive objective is averaged over eligible positive-class anchors
that have at least one other positive example in the batch. If a batch
contains no eligible positive anchor, it is skipped during contrastive
optimization.

\paragraph{MLM Baseline.}
For the Masked Language Modeling (MLM) baseline, we use a standard BERT-style
token masking objective. During pre-training, 15\% of eligible non-padding
clinical tokens in each patient trajectory are randomly selected for
prediction. Among the selected tokens, 80\% are replaced with a special
\texttt{[MASK]} token, 10\% are replaced with a randomly sampled clinical
token, and 10\% are left unchanged. A linear prediction head maps the
256-dimensional token representation at each selected position to logits over
the model vocabulary, and cross-entropy loss is computed only at the selected
positions.

To isolate the effect of the pre-training objective, the MLM baseline uses the
same NHIRD patient subset, encoder architecture, number of training
epochs, batch size, and optimizer settings as the contrastive variants. It is
trained for 10 epochs with batch size 128 using AdamW, a learning rate of
$1\times10^{-4}$, and weight decay $1\times10^{-4}$.

\paragraph{Contrastive Baseline.}
Standard SupCon uses the same encoder architecture, temperature
($\tau=0.07$), optimizer settings, batch size, and number of pre-training
epochs as Asymmetric SupCon, differing only in the contrastive objective.
Specifically, every sample acts as an anchor, and all other samples in the
batch sharing its class label form its positive set. Consequently,
negative-class samples also participate in explicit same-label attraction,
whereas Asymmetric SupCon removes this negative--negative attractive term.

\paragraph{Supervised BCE Source-Transfer Baseline.}
To distinguish the benefit of task-specific source-domain labels from the
contrastive objective, we additionally pre-train the same temporal encoder on
NHIRD using direct supervised binary classification. A two-layer
classification head with the same architecture used for downstream
fine-tuning is attached to the pooled patient representation, and the model is
optimized using weighted binary cross-entropy with a positive-class weight of
5.0. The baseline uses the same 400K-patient NHIRD subset, task labels, number
of training epochs, batch size, AdamW optimizer, learning rate
($1\times10^{-4}$), and weight decay ($1\times10^{-4}$) as the controlled
pre-training ablation. The checkpoint with the lowest validation BCE loss is
retained.

After source-domain training, the NHIRD classification head is discarded.
Only the resulting encoder is transferred to the target dataset, where a new
classification head is initialized and trained using the same downstream
adaptation protocol as the other pre-trained variants.

\paragraph{Downstream Fine-Tuning.}
Downstream hyperparameters were selected based on validation performance;
the selected configuration used for the reported experiments is described
below.

For binary downstream tasks, a two-layer classification head is attached to
the 256-dimensional patient representation. The head consists of a
256-to-128 linear layer, ReLU activation, dropout of 0.3, and a final scalar
output layer. Fine-tuning uses weighted binary cross-entropy with a positive
class weight of 5.0. We optimize with AdamW using learning rates of
$1\times10^{-6}$ for the encoder and $1\times10^{-4}$ for the classifier,
with weight decay $1\times10^{-4}$ and cosine annealing over the training
schedule. The encoder is frozen for the first three epochs and then jointly
fine-tuned with the classifier. Unless otherwise specified, downstream models
are trained for 15 epochs with batch size 64. The checkpoint achieving the
highest validation AUROC is retained.

\paragraph{Evaluation.}
AUROC and AUPRC are computed from predicted probabilities. F1-score is computed
using a fixed probability threshold of 0.5. Final metrics are evaluated on the
held-out test split using the checkpoint selected on validation AUROC.

\section{Semantic Code Mapping Analysis}
\label{appendix:mapping}

To characterize the semantic alignment pipeline, we compare
Qwen3-Embedding-8B \cite{zhang2025qwen3embedding},
SFR-Embedding-Mistral \cite{SFRAIResearch2024},
Linq-Embed-Mistral \cite{choi2024linq}, and
multilingual-e5-large-instruct \cite{wang2024multilingual}
across cosine-similarity thresholds
$\tau_{\mathrm{sim}}\in\{0.7,0.8,0.9\}$.
Across configurations, direct source--target mappings present in our
preprocessing resources are retained. For concepts requiring semantic
alignment, we retrieve the top-ranked compatible NHIRD concept using the specified embedding model and discard
embedding-derived mappings whose cosine similarity falls below the corresponding
threshold.

We additionally characterize target-side mapping collisions. For each distinct
NHIRD target code within a mapping type reached by the mapping procedure, we count the number of
distinct external source concepts mapped to it. A target concept is considered
collision-free when it receives a mapping from exactly one source concept.
We report the collision-free target mapping rate,

\[
\text{Collision-Free Target Rate}
=
\frac{
\#\text{NHIRD target concepts receiving exactly one source mapping}
}{
\#\text{NHIRD target concepts receiving at least one source mapping}
}.
\]

The numerator and denominator are reported in parentheses in
Table~\ref{tab:ablation_mapping}. This statistic measures structural mapping
ambiguity rather than semantic mapping correctness.

Increasing the similarity threshold generally reduces the number of retained
source--target mappings and often increases the collision-free target rate,
although the trend varies across ontologies.
For example, using \texttt{Qwen3-Embedding-8B}, the collision-free target rate
for SNOMED increases from 62.64\% at $\tau_{\mathrm{sim}}=0.7$ to 72.12\% at
$\tau_{\mathrm{sim}}=0.8$ and 92.88\% at $\tau_{\mathrm{sim}}=0.9$, while the
number of distinct NHIRD target concepts reached decreases from 7,838 to 6,463
and then to 3,329.

We use \texttt{Qwen3-Embedding-8B} with
$\tau_{\mathrm{sim}}=0.8$ in the main experiments as a practical operating
point between the breadth of the mapped target vocabulary and target-side
mapping collisions. This mapping procedure serves as an engineering component
for cross-ontology evaluation rather than the primary methodological
contribution of the work.

\begin{table}[H]\centering
\caption{
Collision-free target mapping rate (\%) at different cosine-similarity
thresholds. Parentheses report the number of distinct NHIRD target concepts
receiving exactly one source mapping over the total number of distinct NHIRD
target concepts receiving at least one source mapping.
This statistic measures target-side mapping collisions rather than semantic
mapping correctness. A dash indicates that no mapping survived the threshold.
$^\dagger$ denotes the configuration selected for the main experiments.
}
\label{tab:ablation_mapping}
\scriptsize
\renewcommand{\arraystretch}{1.05}
\setlength{\tabcolsep}{2.6pt}
\resizebox{\textwidth}{!}{
\begin{tabular}{llccccccc}
\toprule
\textbf{Threshold} & \textbf{Model}
& \textbf{SNOMED}
& \textbf{HCPCS}
& \textbf{RxNorm}
& \textbf{CPT-4}
& \textbf{LOINC}
& \textbf{ICD-10-PCS}
& \textbf{ICD-O-3} \\
\midrule

\multirow{4}{*}{$\tau_{\mathrm{sim}}=0.7$}
& \texttt{Qwen3-Embedding-8B}
& 62.64\% (4910/7838)
& 88.10\% (37/42)
& 50.85\% (1137/2236)
& 27.19\% (143/526)
& 33.75\% (243/720)
& 70.59\% (24/34)
& 95.24\% (20/21) \\

& \texttt{SFR-Embedding-Mistral}
& 60.18\% (4789/7958)
& 82.00\% (41/50)
& 55.95\% (1449/2590)
& 30.50\% (212/695)
& 35.67\% (305/855)
& 71.43\% (25/35)
& 90.00\% (18/20) \\

& \texttt{Linq-Embed-Mistral}
& 66.32\% (4780/7207)
& 93.33\% (14/15)
& 62.44\% (1350/2162)
& 26.85\% (109/406)
& 41.68\% (228/547)
& 68.75\% (11/16)
& 95.00\% (19/20) \\

& \texttt{multilingual-e5-large-instruct}
& 57.69\% (4386/7603)
& 87.27\% (48/55)
& 57.61\% (1446/2510)
& 29.69\% (198/667)
& 42.45\% (346/815)
& 75.00\% (27/36)
& 90.00\% (18/20) \\

\midrule

\multirow{4}{*}{$\tau_{\mathrm{sim}}=0.8$}
& \texttt{Qwen3-Embedding-8B}$^\dagger$
& 72.12\% (4661/6463)
& 90.91\% (10/11)
& 54.13\% (1010/1866)
& 41.62\% (82/197)
& 32.39\% (161/497)
& 71.43\% (15/21)
& 95.24\% (20/21) \\

& \texttt{SFR-Embedding-Mistral}
& 61.82\% (4766/7709)
& 90.00\% (27/30)
& 57.40\% (1287/2242)
& 27.69\% (167/603)
& 33.33\% (250/750)
& 67.74\% (21/31)
& 90.00\% (18/20) \\

& \texttt{Linq-Embed-Mistral}
& 78.60\% (4423/5627)
& 100.00\% (1/1)
& 89.25\% (465/521)
& 45.24\% (38/84)
& 67.71\% (130/192)
& 0.00\% (0/2)
& 95.00\% (19/20) \\

& \texttt{multilingual-e5-large-instruct}
& 57.23\% (4315/7540)
& 87.27\% (48/55)
& 56.54\% (1396/2469)
& 29.43\% (196/666)
& 42.24\% (340/805)
& 75.00\% (27/36)
& 90.00\% (18/20) \\

\midrule

\multirow{4}{*}{$\tau_{\mathrm{sim}}=0.9$}
& \texttt{Qwen3-Embedding-8B}
& 92.88\% (3092/3329)
& 100.00\% (1/1)
& 87.79\% (115/131)
& 66.67\% (2/3)
& 41.25\% (66/160)
& --
& 95.00\% (19/20) \\

& \texttt{SFR-Embedding-Mistral}
& 77.33\% (4514/5837)
& 100.00\% (2/2)
& 92.05\% (81/88)
& 30.82\% (49/159)
& 65.18\% (146/224)
& --
& 90.00\% (18/20) \\

& \texttt{Linq-Embed-Mistral}
& 96.60\% (2873/2974)
& --
& --
& 66.67\% (2/3)
& 100.00\% (18/18)
& --
& 100.00\% (17/17) \\

& \texttt{multilingual-e5-large-instruct}
& 57.94\% (4354/7515)
& 90.91\% (30/33)
& 57.37\% (1335/2327)
& 26.93\% (164/609)
& 42.52\% (304/715)
& 74.29\% (26/35)
& 90.00\% (18/20) \\

\bottomrule
\end{tabular}
}
\end{table}

\section{Full EHRSHOT Results}

\begin{table}[H]
\centering
\caption{
Few-shot performance (AUPRC) on the EHRSHOT benchmark. Our results are averaged over three downstream random seeds; baseline results
are from \cite{wornow2023ehrshot}.
Performance is evaluated using all available training data (\textit{All})
and varying numbers of labeled examples per class ($k$).
Our model denotes the task-specific NHIRD-pretrained Asymmetric SupCon
encoder followed by downstream fine-tuning. Best results are bolded and second-best results are underlined.
}
\label{tab:ehrshot_full}

\scriptsize
\renewcommand{\arraystretch}{1.08}
\setlength{\tabcolsep}{2.8pt}

\resizebox{\textwidth}{!}{
\begin{tabular}{llcccccccccccc}
\toprule
\textbf{Task} & \textbf{Model}
& \textbf{All}
& \textbf{1}
& \textbf{2}
& \textbf{4}
& \textbf{8}
& \textbf{12}
& \textbf{16}
& \textbf{24}
& \textbf{32}
& \textbf{48}
& \textbf{64}
& \textbf{128} \\
\midrule

\multirow{5}{*}{Long LOS}
& CLMBR
& \textbf{0.576}
& \underline{0.282}
& \underline{0.304}
& \textbf{0.309}
& \underline{0.315}
& \underline{0.335}
& \underline{0.337}
& \underline{0.358}
& \underline{0.364}
& \underline{0.410}
& \textbf{0.439}
& \textbf{0.473} \\

& GBM
& \underline{0.532}
& 0.255 & 0.253 & 0.265 & 0.295 & 0.292
& 0.296 & 0.305 & 0.323 & 0.349 & 0.380 & 0.403 \\

& Logistic Regression
& 0.384
& 0.237 & 0.251 & 0.283 & 0.283 & 0.283
& 0.294 & 0.308 & 0.311 & 0.346 & 0.340 & 0.365 \\

& Random Forest
& 0.483
& 0.249 & 0.250 & 0.279 & 0.284 & 0.312
& 0.324 & 0.311 & 0.334 & 0.347 & 0.383 & 0.424 \\

& \textbf{AsymSupCon (Ours)}
& 0.517
& \textbf{0.298}
& \textbf{0.306}
& \underline{0.295}
& \textbf{0.333}
& \textbf{0.337}
& \textbf{0.355}
& \textbf{0.387}
& \textbf{0.399}
& \textbf{0.415}
& \underline{0.431}
& \underline{0.450} \\

\midrule

\multirow{5}{*}{ICU Admission}
& CLMBR
& \textbf{0.324}
& \textbf{0.093}
& \textbf{0.095}
& \textbf{0.108}
& \textbf{0.142}
& \textbf{0.152}
& \textbf{0.181}
& \textbf{0.195}
& \textbf{0.215}
& \textbf{0.211}
& \textbf{0.209}
& \textbf{0.256} \\

& GBM
& \underline{0.218}
& 0.046 & 0.046 & 0.049 & 0.064 & 0.065
& 0.073 & 0.079 & 0.109 & 0.113
& \underline{0.129} & 0.165 \\

& Logistic Regression
& 0.094
& \underline{0.056}
& 0.053 & 0.054 & 0.067 & 0.076
& 0.079 & 0.089 & 0.093 & 0.081 & 0.078 & 0.097 \\

& Random Forest
& 0.089
& 0.055 & 0.048 & 0.056 & 0.069 & 0.070
& 0.086 & 0.095
& \underline{0.127}
& \underline{0.115}
& 0.116
& \underline{0.190} \\

& \textbf{AsymSupCon (Ours)}
& 0.138
& 0.055
& \underline{0.057}
& \underline{0.068}
& \underline{0.080}
& \underline{0.082}
& \underline{0.090}
& \underline{0.096}
& 0.112 & 0.105 & 0.108 & 0.129 \\

\midrule

\multirow{5}{*}{30-Day Readmission}
& CLMBR
& \textbf{0.412}
& 0.135
& \textbf{0.182}
& \textbf{0.253}
& \underline{0.237}
& \textbf{0.267}
& \textbf{0.297}
& \textbf{0.317}
& \textbf{0.329}
& \textbf{0.364}
& \textbf{0.368}
& \textbf{0.350} \\

& GBM
& 0.327
& 0.119 & 0.133 & 0.134 & 0.189 & 0.169
& 0.177 & 0.215 & 0.212 & 0.228 & 0.245 & 0.276 \\

& Logistic Regression
& 0.258
& \underline{0.150}
& 0.162 & 0.190 & 0.201 & 0.196
& 0.194 & 0.191 & 0.195 & 0.207 & 0.217 & 0.229 \\

& Random Forest
& 0.335
& 0.145 & 0.142 & 0.147 & 0.175 & 0.169
& 0.205 & 0.213 & 0.206 & 0.240 & 0.240 & 0.267 \\

& \textbf{AsymSupCon (Ours)}
& \underline{0.343}
& \textbf{0.172}
& \underline{0.177}
& \underline{0.226}
& \textbf{0.238}
& \underline{0.234}
& \underline{0.243}
& \underline{0.255}
& \underline{0.264}
& \underline{0.271}
& \underline{0.278}
& \underline{0.289} \\

\midrule

\multirow{5}{*}{Acute MI}
& CLMBR
& 0.183
& \underline{0.086}
& \underline{0.093}
& \underline{0.120}
& \underline{0.096}
& \underline{0.099}
& \underline{0.118}
& \underline{0.137}
& \underline{0.159}
& \underline{0.138}
& \underline{0.154}
& \underline{0.158} \\

& GBM
& \underline{0.184}
& 0.066 & 0.073 & 0.081 & 0.087 & 0.084
& 0.107 & 0.122 & 0.117 & 0.134 & 0.152
& \underline{0.158} \\

& Logistic Regression
& 0.135
& 0.082 & 0.087 & 0.098 & 0.092 & 0.093
& 0.092 & 0.106 & 0.103 & 0.110 & 0.111 & 0.116 \\

& Random Forest
& 0.165
& 0.083 & 0.082 & 0.088
& \underline{0.096}
& 0.090 & 0.103 & 0.111 & 0.122 & 0.128 & 0.143 & 0.152 \\

& \textbf{AsymSupCon (Ours)}
& \textbf{0.191}
& \textbf{0.106}
& \textbf{0.094}
& \textbf{0.135}
& \textbf{0.135}
& \textbf{0.128}
& \textbf{0.153}
& \textbf{0.163}
& \textbf{0.179}
& \textbf{0.175}
& \textbf{0.178}
& \textbf{0.178} \\

\midrule

\multirow{5}{*}{Pancreatic Cancer}
& CLMBR
& 0.252
& 0.047
& \underline{0.059}
& \textbf{0.087}
& \underline{0.080}
& \underline{0.102}
& \underline{0.123}
& \underline{0.123}
& \underline{0.134}
& 0.137
& 0.151
& 0.177 \\

& GBM
& \underline{0.372}
& 0.027 & 0.028 & 0.040 & 0.068 & 0.094
& 0.071 & 0.080 & 0.128
& \underline{0.177}
& \underline{0.181}
& \textbf{0.243} \\

& Logistic Regression
& 0.157
& \underline{0.058}
& 0.043 & 0.050 & 0.051 & 0.056
& 0.063 & 0.072 & 0.067 & 0.085 & 0.115 & 0.122 \\

& Random Forest
& \textbf{0.472}
& 0.028 & 0.040 & 0.049 & 0.061 & 0.069
& 0.086 & 0.103 & 0.113 & 0.136 & 0.153 & 0.172 \\

& \textbf{AsymSupCon (Ours)}
& 0.305
& \textbf{0.074}
& \textbf{0.067}
& \underline{0.074}
& \textbf{0.095}
& \textbf{0.131}
& \textbf{0.151}
& \textbf{0.150}
& \textbf{0.166}
& \textbf{0.179}
& \textbf{0.187}
& \underline{0.229} \\

\midrule

\multirow{5}{*}{Lupus}
& CLMBR
& 0.024
& \underline{0.012}
& 0.011
& \underline{0.012}
& \underline{0.015}
& 0.015
& 0.016
& 0.018
& 0.026
& 0.020
& 0.018
& 0.018 \\

& GBM
& \textbf{0.179}
& 0.009 & 0.010 & 0.011
& \underline{0.015}
& \underline{0.030}
& \underline{0.017}
& \underline{0.029}
& \underline{0.036}
& \underline{0.058}
& \underline{0.061}
& \underline{0.079} \\

& Logistic Regression
& \underline{0.093}
& 0.011
& \underline{0.012}
& 0.010 & 0.012 & 0.012 & 0.015 & 0.013
& 0.015 & 0.016 & 0.018 & 0.019 \\

& Random Forest
& 0.016
& 0.009
& \underline{0.012}
& 0.010 & 0.012 & 0.014
& 0.016 & 0.025 & 0.022 & 0.036 & 0.055 & 0.049 \\

& \textbf{AsymSupCon (Ours)}
& \underline{0.093}
& \textbf{0.022}
& \textbf{0.014}
& \textbf{0.028}
& \textbf{0.024}
& \textbf{0.038}
& \textbf{0.036}
& \textbf{0.070}
& \textbf{0.066}
& \textbf{0.084}
& \textbf{0.085}
& \textbf{0.084} \\

\midrule

\multirow{5}{*}{Hyperlipidemia}
& CLMBR
& 0.225
& 0.142 & 0.149 & 0.161
& \underline{0.180}
& \underline{0.175}
& 0.179
& \underline{0.193}
& 0.191
& 0.181
& 0.198
& 0.214 \\

& GBM
& \underline{0.258}
& 0.126 & 0.130 & 0.142 & 0.158 & 0.152
& 0.157 & 0.165 & 0.202 & 0.184
& \underline{0.210}
& 0.222 \\

& Logistic Regression
& 0.247
& \underline{0.144}
& \underline{0.166}
& \underline{0.164}
& 0.169 & 0.174
& \underline{0.183}
& 0.188
& \underline{0.212}
& \underline{0.203}
& 0.208
& \underline{0.244} \\

& Random Forest
& 0.193
& 0.138 & 0.143 & 0.151 & 0.161 & 0.149
& 0.160 & 0.181 & 0.195 & 0.193 & 0.186 & 0.227 \\

& \textbf{AsymSupCon (Ours)}
& \textbf{0.297}
& \textbf{0.146}
& \textbf{0.179}
& \textbf{0.194}
& \textbf{0.207}
& \textbf{0.224}
& \textbf{0.234}
& \textbf{0.240}
& \textbf{0.238}
& \textbf{0.240}
& \textbf{0.248}
& \textbf{0.265} \\

\midrule

\multirow{5}{*}{Hypertension}
& CLMBR
& \underline{0.258}
& \textbf{0.151}
& \underline{0.189}
& \underline{0.188}
& \underline{0.205}
& \underline{0.213}
& \underline{0.224}
& \underline{0.222}
& \underline{0.220}
& \underline{0.226}
& \underline{0.233}
& \underline{0.243} \\

& GBM
& 0.221
& 0.127 & 0.129 & 0.140 & 0.138 & 0.148
& 0.152 & 0.175 & 0.165 & 0.198 & 0.209 & 0.206 \\

& Logistic Regression
& 0.212
& 0.138 & 0.151 & 0.158 & 0.147 & 0.160
& 0.175 & 0.176 & 0.169 & 0.182 & 0.188 & 0.191 \\

& Random Forest
& 0.226
& 0.133 & 0.142 & 0.146 & 0.159 & 0.164
& 0.175 & 0.175 & 0.167 & 0.190 & 0.208 & 0.203 \\

& \textbf{AsymSupCon (Ours)}
& \textbf{0.295}
& \underline{0.139}
& \textbf{0.204}
& \textbf{0.242}
& \textbf{0.254}
& \textbf{0.267}
& \textbf{0.266}
& \textbf{0.258}
& \textbf{0.256}
& \textbf{0.269}
& \textbf{0.279}
& \textbf{0.280} \\

\bottomrule
\end{tabular}
}

\end{table}

\end{document}